\documentclass{article}
\usepackage{amssymb}
\usepackage{amsmath}
\usepackage{graphicx}
\usepackage{subcaption}
\usepackage{booktabs}
\usepackage{caption}
\usepackage{overpic}
\usepackage{xcolor}
\usepackage{enumitem}
\usepackage[colorlinks=true, urlcolor=blue,]{hyperref}
\usepackage{acronym}
\usepackage{array}
\usepackage{wrapfig}
\acrodef{qp}[QP]{Quadratic Programming}
\acrodef{dofs}[DoFs]{Degrees of Freedom}
\acrodef{ros}[ROS]{Robot Operating System}
\acrodef{com}[CoM]{center of mass}
\acrodef{lipm}[LIPM]{Linear Inverted Pendulum Model}
\acrodef{ocp}[OCP]{Optimal Control Problem}
\acrodef{mpc}[MPC]{Model Predictive Control}
\acrodef{wbc}[WBC]{Whole-Body Control}
\acrodef{cvae}[CVAE]{Conditional Variational Autoencoder }
\acrodef{ik}[IK]{Inverse Kinematics}
\acrodef{rl}[RL]{Reinforcement Learning}
\acrodef{il}[IL]{Imitation Learning}
\acrodef{to}[TO]{Trajectory Optimization}
\acrodef{imu}[IMU]{Inertial Measurement Unit}
\acrodef{zmp}[ZMP]{Zero Moment Point}
\acrodef{mocap}[MoCap]{motion capture}
\acrodef{gmr}[GMR]{General Motion Retargeting}
\acrodef{ppo}[PPO]{Proximal Policy Optimization}

\usepackage[preprint]{corl_2026} 

\title{T-GMP: Terrain-conditioned Generative Motion Priors for Versatile and Natural Humanoid Locomotion}

\author{
\textbf{Junhong Guo}$^{1,2,*}$\quad
\textbf{Hao Hu}$^{1,*}$\quad
\textbf{Chen Chen}$^{1}$\quad
\textbf{Haoxuan Han}$^{1}$\quad
\textbf{Linao Gong}$^{1}$\\
\textbf{Xin Yang}$^{1}$\quad
\textbf{Zhicheng He}$^{1,2}$\quad
\textbf{Yao Su}$^{2}$\quad
\textbf{Fenghua He}$^{1,\dagger}$ \\
$^1$ Harbin Institute of Technology\\
$^2$ Leju Robotics\\
{\footnotesize
$^{*}$ Equal contribution; order decided by coin toss.\quad
$^{\dagger}$ Corresponding author.
}
}

\begin{document}
\maketitle


\vspace{-10mm}
\begin{figure}[h]
    \centering
    \includegraphics[width=1.0\textwidth]{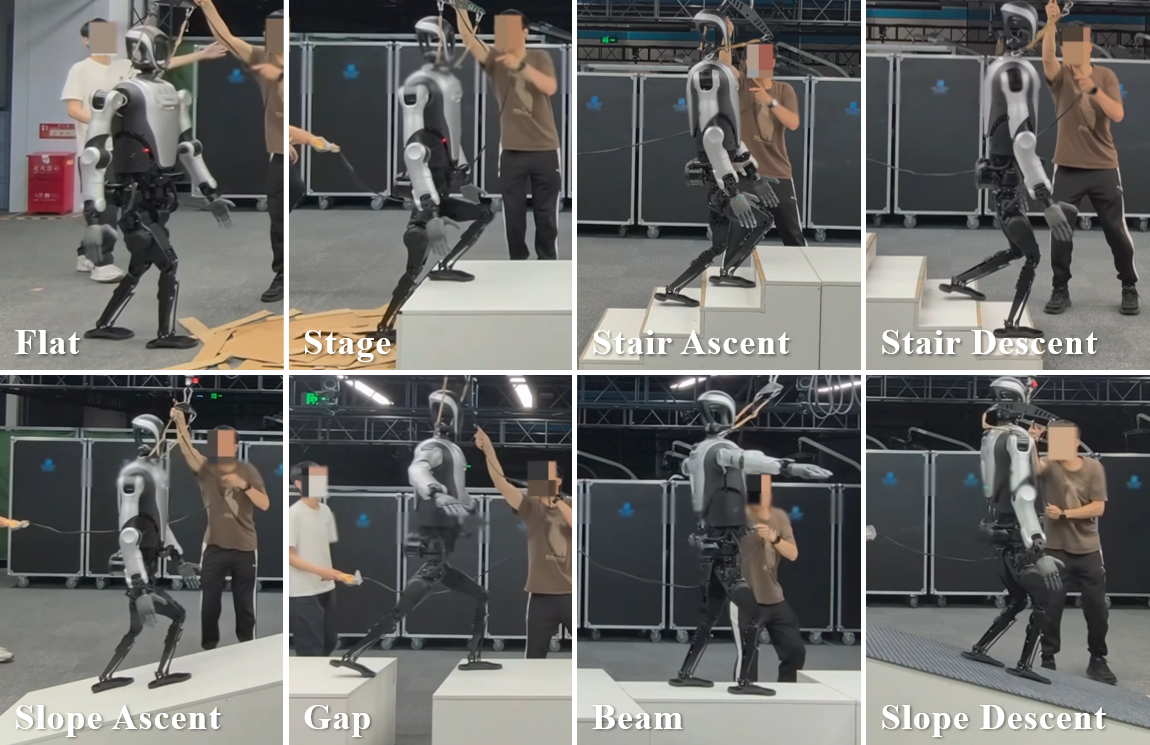}
    \caption{T-GMP enables robots to exhibit versatile and natural behaviors across diverse terrains. Trained only once, the robot learns terrain-adaptive whole-body coordination strategies, including natural arm swinging during walking, lowering the \ac{com} on stairs and slopes, and extending its arms for balance when crossing gaps or narrow beams. Videos are available on our project website: \href{https://t-gmp.github.io/T-GMP}{https://t-gmp.github.io}.
    }
    \label{fig:cover}
\end{figure}

\vspace{-4mm}
\begin{abstract}
Achieving both anthropomorphic naturalness and rich motion diversity during terrain traversal remains a fundamental challenge in humanoid locomotion. Existing reinforcement learning approaches typically rely on fixed motion priors, limiting their adaptability to varying environments. We propose Terrain-conditioned Generative Motion Priors (T-GMP), a module that captures a terrain-conditioned latent motion manifold from a few expert state-terrain demonstrations. The learned priors enable smooth style transitions, facilitating a unified policy that adapts to terrain variations. We integrate T-GMP into an adversarial learning pipeline, where a discriminator dynamically modulates naturalness constraints conditioned on local terrain features, guiding the generation of versatile and human-like motions. We further introduce a Foothold Penalty to promote safe foot placement on challenging terrains. Experimental results demonstrate that T-GMP outperforms existing baselines in motion naturalness, motion diversity, and traversal success rates, while preserving physically coordinated motions.
\end{abstract}

\vspace{-2mm}

\section{Introduction}

The pursuit of versatile and human-like locomotion in humanoid robots is a fundamental goal in robotics, bridging the gap between mechanical efficiency and biological elegance. While \ac{rl} approaches have achieved remarkable success in enabling robots to locomote robustly~\cite{Vanilla, HumanoidLocomotionReinforcement, Li2024ReinforcementLF, HumanoidMamba} and traverse challenging terrains~\cite{ANYmalParkour, Radosavovic2024LearningHL, PerceptiveHumanoidLoco, Gallant}, the resulting motions often exhibit ``robotic'' artifacts—stiffness, unnatural joint configurations, or a lack of biological plausibility. This is primarily because hand-crafted reward functions are often insufficient to capture the nuanced dynamics of human balance, especially when the robot must simultaneously optimize task objectives (e.g., velocity tracking) and stylistic constraints across diverse terrains.

To improve motion quality, mimic-based approaches~\cite{DeepMimic, BeyondMimic, VideoMimic, genMimic} have been widely adopted to distill ``naturalness'' and ``agility'' from \ac{mocap} data into policies, enabling robots to demonstrate highly dynamic skills such as jumping, dancing, and backflipping. However, when a robot encounters complex terrains, fixed and terrain-agnostic motion priors penalize the necessary deviations required for physical stability. This creates a fundamental conflict between task completion and style preservation, often leading to stiff gaits or failure under varying environments.

To address these limitations, we propose T-GMP, a module designed for adaptive and anthropomorphic humanoid locomotion. Our core insight is that human motion is intrinsically linked to the environment; therefore, a motion prior should not be a static template but a dynamic manifold conditioned on terrain features. By utilizing a \acf{cvae}, T-GMP learns a terrain-conditioned latent motion manifold from a few expert state-terrain demonstrations. This enables the robot to learn how anthropomorphic motion styles adapt to different environments and achieve smooth style transitions across varying terrains.

Furthermore, we integrate T-GMP into a unified adversarial \ac{rl} framework, where the proposed terrain-conditioned discriminator injects terrain-aware anthropomorphic motion-style constraints into policy learning. To improve foothold stability on complex terrains, we further introduce a Foothold Penalty that explicitly regularizes foot-terrain contact quality, thereby enhancing locomotion robustness. Experimental results demonstrate that the proposed method generates coordinated and anthropomorphic whole-body behaviors across diverse terrains, while significantly outperforming baselines in motion smoothness, motion diversity and traversal success rates.

Overall, our contributions can be summarized as follows:
\begin{enumerate}[leftmargin=*, noitemsep]
\item[1)] We propose T-GMP, a module that learns a terrain-conditioned latent motion manifold from a few expert state-terrain demonstrations, enabling the generation of terrain-adaptive motion priors.
\item[2)] By introducing a terrain-conditioned discriminator, we integrate T-GMP into a unified adversarial \ac{rl} pipeline that jointly optimizes task objectives and motion style constraints.
\item[3)] We conduct real-world experiments on a full-size humanoid robot across eight terrains, demonstrating diverse, coordinated, and anthropomorphic whole-body behaviors. Simulation experiments on multiple unseen terrains further demonstrate the generalization of the learned policy.
\end{enumerate}

\section{Related Works}

\paragraph{Legged Locomotion} Recent advancements in humanoid locomotion have seen a transition from proprioceptive-only methods to perception-aware approaches. While proprioceptive-only methods~\cite{QuadrupedalLoco, RMA, HIM, DreamWaq, HumanoidLocomotionReinforcement} demonstrate impressive robustness through proprioceptive feedback, they are limited by a reactive control loop that may cause hardware damage during abrupt terrain changes. Conversely, perception-aware policies utilize depth sensors~\cite{DreamWaq++, EgocentricVisionLeggedLoco, RobotParkourLearning, HumanoidParkourLearning, HikingInTheWild, Miki, ExtremeParkour} or LiDAR~\cite{CNNLocomotion, PIM, Beamdojo, DTC, AttentionBasedMapEncoding4Humanoid, AME2} to construct egocentric terrain representations and enable proactive maneuvers. However, existing perception-driven methods predominantly focus on lower-limb stability and gait scheduling, frequently overlooking the whole-body coordination potential inherent in the humanoid form. Our work addresses this gap by phrasing locomotion as a terrain-conditioned whole-body coordination task. By integrating perception with generative motion priors, our framework enables the robot to autonomously recruit its entire skeletal structure, including arms and torso, to maintain stability over challenging terrains.

\paragraph{Generative Motion Priors} Leveraging human motion priors to enhance anthropomorphic locomotion has become a prevailing direction in recent robotics research. Mimic-based approaches~\cite{DeepMimic, vmp, Twist2} directly track human \ac{mocap} data, enabling robots to exhibit natural and versatile skills. However, such methods typically rely on fixed reference trajectories, which limit their ability to generalize to multi-terrain traversal scenarios with complex and varying conditions. To improve generalization, subsequent studies abstract human motion priors into transferable motion styles that guide policies to produce human-like behaviors~\cite{amp, ASE, C-ASE, BeyondMimic, PhysHSI, HIL}. Although these methods partially alleviate the dependence on a single reference trajectory, the underlying motion priors are still fixed, making it difficult to adapt to diverse terrains. More recently, generative motion priors~\cite{naturalLoco, RuN, DreamPolicy, zhang2026learningwholebodyhumanoidlocomotion} have been introduced to model richer motion distributions, enabling robots to exhibit natural motion styles and smooth style transitions. We further integrate generative motion priors with terrain perception, generating motion priors conditioned on local terrain features during the locomotion policy training process. Moreover, we employ a terrain-conditioned discriminator to distill the motion priors into the policy during training, enabling natural and smooth anthropomorphic locomotion across various terrains.

\section{Terrain-conditioned Generative Motion Priors}

\subsection{Data Collection}

We develop a data collection pipeline to construct a natural, diverse, and physically feasible expert state–terrain dataset. For each terrain $i$ in the terrain set $\mathcal I$, we first set up the corresponding terrain in a motion-capture studio and record human \ac{mocap} data, resulting in a terrain-specific MoCap dataset $\{\mathcal D_i^{\text{mocap}}\}_{i\in\mathcal I}$. We then retarget these motions to the robot joint space using GMR~\cite{GMR}, yielding a set of natural and diverse robot motion references $\{\mathcal D_i^{\text{robot}}\}_{i\in\mathcal I}$. However, the retargeted motions may contain physically infeasible artifacts, such as floating feet and excessive body tilt. To mitigate these artifacts while facilitating the collection of state-terrain data, we use $\mathcal D_i^{\text{robot}}$ as motion priors and train a privileged expert policy $\pi_i^{\text{priv}}$ separately on each terrain $i$. 

We roll out the trained policy $\pi_i^{\text{priv}}$ to collect state–terrain sequences $\{(s_t,h_t)\}_{t=1}^L$, where $L$ denotes the trajectory length. The expert state is defined as:
\begin{equation}
s_t=[q_t,\dot q_t,p_t],
\end{equation}
where $q_t$ and $\dot q_t$ denote joint positions and velocities, and $p_t$ denotes end-effector positions relative to the root frame. The perceptual observation $h_t$ is the local height map. By aggregating the state–terrain sequences collected across all terrains, we construct a high-quality expert dataset $\mathcal D$ that captures natural and diverse motions while satisfying the robot’s physical constraints. Implementation details of privileged expert training and state–terrain data collection are provided in Section~\ref{app:data collection}.

\begin{figure}[t]
    \centering
    \includegraphics[width=1.0\textwidth]{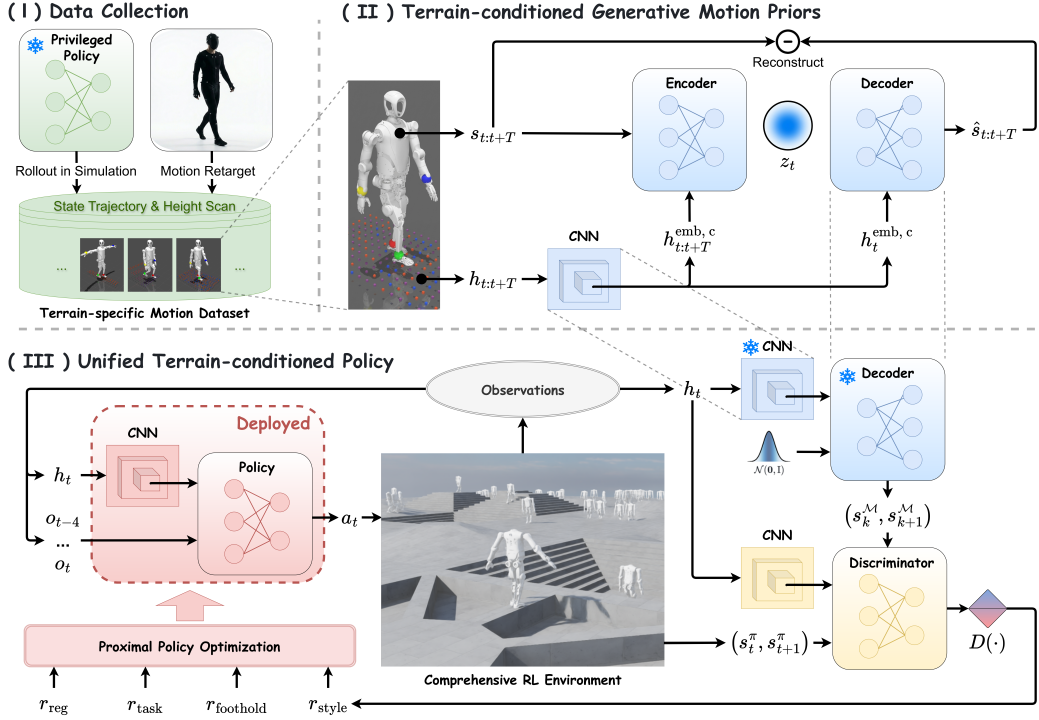}
    \caption{\textbf{Method Overview}. The overall learning pipeline consists of three parts: (I) collecting expert data using privileged policies, (II) training T-GMP using a \ac{cvae}, and (III) training a unified \ac{rl} policy with a terrain-conditioned discriminator. The terrain representation (height map) serves as a conditioning variable across all modules.
    }
    \label{fig:overview}
\end{figure}

\subsection{Terrain-conditioned Generative Motion Priors}
\label{sec:motion prior generator}

We adopt a \ac{cvae} to model multi-style motion distributions, and use its decoder as T-GMP. To explicitly incorporate terrain-conditioned information, we employ a two-layer CNN $f_{\text{cnn}}^\text{c}$ to extract features from the local height map $h_t$, yielding a terrain embedding $h_t^{\text{emb},~\text{c}}=f_{\text{cnn}}^\text{c}(h_t)$. Building on this representation, we extend the standard $\beta$-VAE~\cite{BetaVAE} to a conditional $\beta$-VAE formulation. The \ac{cvae} decoder conditions on the terrain embedding $h_t^{\text{emb},~\text{c}}$, together with a latent variable $z_t$ sampled from the prior distribution $p(z_t)$, generates expert state trajectory sequences, which exhibit diverse motion styles across different terrains:
\begin{equation}
    \hat s_{t:t+T} = \text{Decoder}(z_t,h_t^{\text{emb},~\text{c}}).
\end{equation}

The overall architecture of the CVAE is illustrated in Figure~\ref{fig:overview} (II). The model is trained by minimizing a weighted sum of a reconstruction loss and a regularization term, defined as:
\begin{equation}
\begin{aligned}
\mathcal L_{\text{cvae}}=&\quad\mathbb E_{(s_{t:t+T},h_{t:t+T})\sim\mathcal D}[\mathbb E_{q_\phi(z_t|s_{t:t+T},h_{t:t+T}^{\text{emb},~\text{c}})}[\|s_{t:t+T}-\hat s_{t:t+T}\|^2_2]\\
&+\beta D_{\text{KL}}(q_\phi(z_t|s_{t:t+T},h_{t:t+T}^{\text{emb},~\text{c}})||p(z_t))],
\end{aligned}
\end{equation}
where $s_{t:t+T}$ and $h_{t:t+T}^{\text{emb},~\text{c}}$ respectively denote a sequence of $T$ consecutive expert states and their corresponding local height map features, and $\beta$ balances reconstruction accuracy and latent space regularization. We adopt a standard normal distribution $\mathcal N(0, I)$ as the prior distribution $p(z_t)$.

During real-world deployment, only the local height map observation at the current time step is available. Therefore, we condition the decoder solely on a single-frame height map $h_t$ to ensure consistency between training and inference. To mitigate error accumulation caused by insufficient conditioning information, we shorten the reconstruction horizon $T=16$, thereby effectively limiting the propagation of long-horizon prediction errors. The CVAE is trained offline and can be reused across downstream training tasks without retraining.

\section{Unified Terrain-conditioned Policy Training}
\subsection{Terrain-conditioned Discriminator}

To enable the robot to exhibit natural and versatile humanoid motions with coordinated whole-body behaviors when traversing diverse terrains, we introduce adversarial motion priors (AMP)~\cite{amp} to endow the policy with anthropomorphic style from the expert dataset. However, when multiple motion priors simultaneously cover the same task, the standard AMP framework may cause distinct motion styles to collapse into indistinguishable behaviors during policy learning. To address this problem, we extend the AMP discriminator to a terrain-conditioned discriminator. Specifically, we employ a five-layer CNN $f_{\text{cnn}}^\text{d}$ to extract terrain features from the local height map $h_t$, and use the resulting terrain embedding $h_t^{\text{emb},~\text{d}}$ as a conditioning input, together with state transitions, to guide the policy toward learning distinct and diverse motion styles under varying terrains.

During discriminator training, we use the decoder trained in Section~\ref{sec:motion prior generator} to generate expert state trajectory sequences $\tau^{\mathcal M}_t=[s^{\mathcal M}_{t},\cdots,s^{\mathcal M}_{t+T}]$ online conditioned on the terrain features, from which expert state transition pairs $(s^{\mathcal M}_{k},s^{\mathcal M}_{k+1})$ are randomly sampled, where $t\le k<t+T$. In parallel, policy-induced state transition pairs $(s^\pi_t,s^\pi_{t+1})$ are obtained from environment rollouts by executing the current policy, where $s^\pi_t$ denotes the policy state at time step $t$ and $s^\pi_{t+1}$ denotes the resulting next state under the policy. 

Unlike the formulation in AMP~\cite{amp}, our terrain-conditioned discriminator $D$ is trained by minimizing the following objective:
\begin{equation}
\begin{aligned}
\underset{D}{\arg\min}\quad&\mathbb E_{d^\mathcal M(s_k,s_{k+1})}[(D(s_k,s_{k+1}|h_t^{\text{emb},~\text{d}})-1)^2]\\
+&\mathbb E_{d^\pi(s_t,s_{t+1})}[(D(s_t,s_{t+1}|h_t^{\text{emb},~\text{d}})+1)^2]\\
+&\frac{\mathbf w^{\text{gp}}}{2} \mathbb E_{d^\mathcal M(s_k,s_{k+1})}[\|\nabla_\phi D(\phi)|_{\phi=(s_k,s_{k+1}|h_t^{\text{emb},~\text{d}})}\|^2],
\end{aligned}
\end{equation}
where $h_t^{\text{emb},~\text{d}}$ represents local height map features extracted by the five-layer CNN $f_{\text{cnn}}^\text{d}$, $d^{\mathcal M}$ and $d^\pi$ respectively represent the decoder-generated expert state transition distribution and the policy-induced state transition distribution, and $\mathbf w^{\text{gp}}$ is the weight of the gradient penalty term.

During policy training, to provide the policy with a clear and stable style guidance signal, we convert the discriminator output into a style reward, which is defined as:
\begin{equation}
r_{\text{amp}} = \max[0,~1-(D(s^\pi_t,s^\pi_{t+1}|h_t^{\text{emb},~\text{d}})-1)^2/4].
\end{equation}
This reward function encourages policy-induced state transitions to align with the terrain-conditioned expert motion styles, thereby guiding the policy to acquire consistent yet terrain-distinctive locomotion behaviors across diverse terrains.

\subsection{Policy Training}

\paragraph{Observations} We define the robot’s observation at time step $t$ as:
\begin{equation}
\mathbf o_t = [\omega_t,\mathbf g_t,q_t,\dot q_t,a_{t-1}],
\end{equation}
where $\omega_t$ denotes the angular velocity of the robot’s root, $\mathbf g_t$ is the gravity direction vector, $q_t$ and $\dot q_t$ respectively represent joint positions and velocities, and $a_{t-1}\in \mathbb R^{26}$ denotes the action executed at the previous time step. To enhance temporal awareness and robustness, we stack the most recent five frames of observations, forming an augmented observation $\mathbf O_t=[\mathbf o_{t-4},\cdots,\mathbf o_{t}]$. In addition, the Critic network has access to privileged information $\mathbf o^{\text{priv}}_t$ during training to improve the accuracy of value estimation. Detailed information about $\mathbf o_t$ and $\mathbf o_t^{\text{priv}}$ is provided in Section~\ref{app:observations}.

\paragraph{Policy} The overall training framework is illustrated in Figure~\ref{fig:overview} (III). To enable the policy to exhibit discriminative motion styles while maintaining strong terrain adaptability across different terrains, we explicitly incorporate terrain perception into the policy network. Specifically, we employ a CNN $f_{\text{cnn}}^\text{p}$ with the same architecture described in Section~\ref{sec:motion prior generator} to extract features from the local height map $h_t$, and concatenate the resulting terrain embedding $h_t^{\text{emb},~\text{p}}$ with the robot’s augmented proprioceptive observation $\mathbf O_t$ and velocity command $c_t$ as input to the policy network. This design allows the policy to: (1) learn differentiated motion styles under varying terrains, and (2) attend to local terrain features, thereby perceiving terrain changes and guiding the robot to make more anticipatory decisions. Detailed policy training configurations are provided in Section~\ref{app:rl learning setup}.

\paragraph{Foothold Penalty} During policy training, the robot may exhibit undesirable foothold behaviors when interacting with terrain edges and foothold contacts, such as kicking stair edges during ascent and edge-sliding during descent, which significantly degrade motion naturalness and physical plausibility. To address this problem, we introduce Foothold Penalty, which explicitly constrains foot-terrain contact quality through ray-cast distance measurements from the toe and sole regions. Specifically, the toe distance is used to penalize toe collisions:
\begin{equation}
r_{\text{toe}}=-\max(0,\,(\frac{d_{\text{toe}}^{\text{threshold}}-d_{\text{toe}}}{d_{\text{toe}}^{\text{threshold}}})^3),
\end{equation}
where $d_{\text{toe}}$ denotes the minimum distance between the toe and the terrain surface, $d_{\text{toe}}^{\text{threshold}}$ denotes the safety distance threshold. The sole distance suppresses edge-support and sliding behaviors:
\begin{equation}
r_{\text{sole}}=-\mathrm{clip}(d_{\text{sole}}-d_{\text{sole}}^{\text{threshold}},\,0,\,1)\cdot \mathbb{I}_{\text{contact}},
\end{equation}
where $d_{\text{sole}}$ denotes the maximum distance between the sole and the ground surface, $d_{\text{sole}}^{\text{threshold}}$ denotes the maximum allowable support distance, and $\mathbb{I}_{\text{contact}}$ denotes a contact indicator. Finally, we introduce the Foothold Penalty: $r_{\text{foothold}}=r_{\text{toe}}+r_{\text{sole}}$. An illustration of the proposed Foothold Penalty is provided in Figure~\ref{fig:foothold penalty}. More reward settings can be found in Section~\ref{app:reward formulation}. 

\section{Experiment}
\label{sec:experiment}

\subsection{Experimental Setup}

We conduct policy training and deployment on a 28-DoF full-size Kuavo humanoid robot. To evaluate the advantages of our method, we consider the following baselines: (1) a vanilla \ac{rl} baseline ~\cite{Vanilla} (\textbf{Baseline RL}); and (2) a standard AMP~\cite{amp} (\textbf{AMP}). To evaluate the contribution of each key component, we consider three ablations: (1) removing the CVAE module and instead randomly sampling expert state transitions from the expert dataset $\mathcal D$ (\textbf{Ours w/o CVAE}); (2) removing the terrain-conditioning embedding $h_t^{\text{emb},~\text{d}}$ from the discriminator (\textbf{Ours w/o Condition}); and (3) removing the Foothold Penalty (\textbf{Ours w/o Foothold}). All policies are trained on the same dataset using 2048 parallel environments on a single NVIDIA RTX 4090 GPU.

\subsection{Motion Diversity}
\label{sec:Motion Diversity}

\begin{figure}[t]
    \centering
    \includegraphics[width=1.0\textwidth]{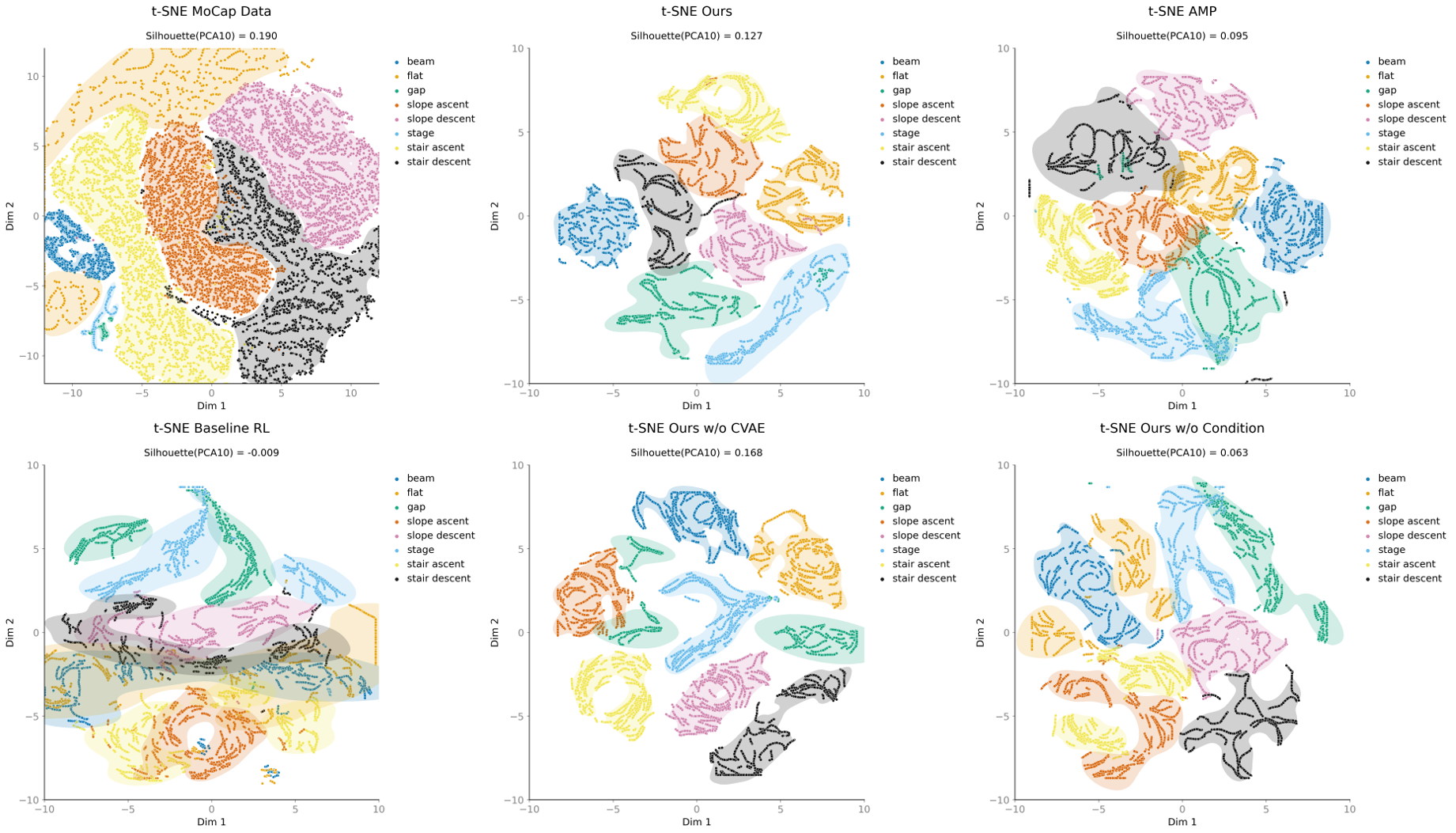}
    \caption{
    t-SNE visualization of MoCap and policy-generated motion distributions across different terrains, with Silhouette Scores computed on 10-dimensional features extracted using Principal Component Analysis (PCA).
    }
    \label{fig:tsne}
\end{figure}

Our simulation (Figure~\ref{fig:unseen terrain}) and real-world (Figure~\ref{fig:all style}) experiments qualitatively demonstrate that the robot exhibits diverse motion styles when traversing different terrains. To quantitatively evaluate the motion style distributions learned by different methods, we collect 700-step joint-position trajectories $\{q_t\}_{t=1}^{700}$ from five methods across all terrains and visualize their distributions using t-SNE~\cite{t-SNE}. As shown in Figure~\ref{fig:tsne}, the comparison methods exhibit less compact motion distributions with significant inter-terrain overlap. In contrast, our method produces highly compact intra-terrain clusters and clearly separable inter-terrain distributions. We further report the Silhouette Score to jointly quantify intra-terrain consistency and inter-terrain separability, where a higher score indicates better cluster separation. Our method achieves a score of 0.127, outperforming AMP (0.095) and all other baselines except Ours w/o CVAE (0.168). Ours w/o CVAE learns disordered styles, thus yielding an anomalously high score, which we will explain in Section~\ref{sec:Motion Style Visualization}.

\subsection{Motion Naturalness}
\label{sec:Motion Naturalness}

\begin{figure}[t]
    \centering
    \includegraphics[width=1.0\textwidth]{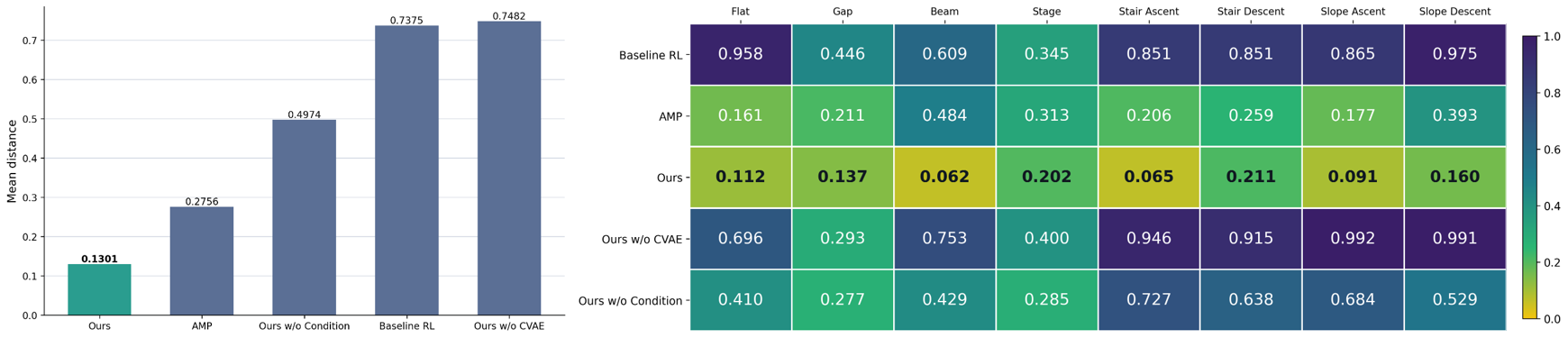}
    \caption{
    MMD between policy-generated and MoCap motion distributions across all terrains. Lower values indicate closer agreement with the MoCap motion distribution.
    }
    \label{fig:mmd}
\end{figure}

\begin{figure}[t]
    \vspace{-6pt}
    \centering
    \includegraphics[width=1.0\textwidth]{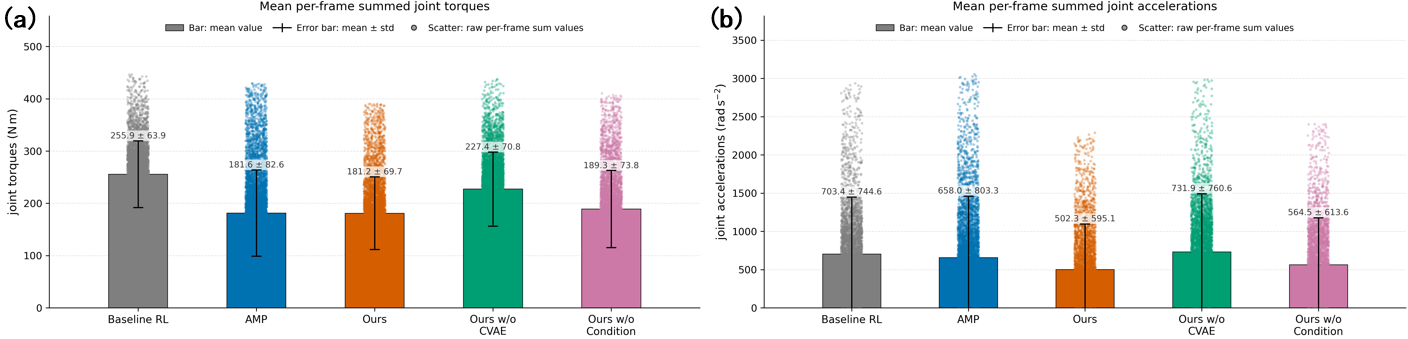}
    \caption{
    Motion smoothness analysis across all terrains. (a) Mean and standard deviation of joint torques. (b) Mean and standard deviation of joint accelerations.
    }
    \label{fig:joint metrics}
\end{figure}

\begin{table}[b]
    \centering
    \vspace{-6pt}
    \caption{
    Terrain-traversal success rates across seven challenging terrains.
    }
    \label{tab:success rate}
    \resizebox{\textwidth}{!}{
    \begin{tabular}{l*{7}{>{\centering\arraybackslash}m{1.2cm}}}
        \toprule
        & \multicolumn{7}{c}{\textbf{Success Rate (\%)}} \\
        \cmidrule(lr){2-8}
        \textbf{Method} & \textbf{Gap} & \textbf{Beam} & \textbf{Stage} & \textbf{Stair Ascent} & \textbf{Stair Descent} & \textbf{Slope Ascent} & \textbf{Slope Descent} \\
        \midrule
        Baseline RL~\cite{Vanilla} & 92.97 & 79.69 & 93.36 & 86.33 & 85.55 & 92.19 & 95.31 \\
        AMP~\cite{amp} & 93.02 & 86.72 & 90.33 & 89.06 & 82.56 & 93.41 & 96.09 \\
        Ours & \textbf{98.83} & \textbf{96.88} & \textbf{96.09} & \textbf{96.48} & \textbf{93.75} & \textbf{99.61} & \textbf{100.00} \\
        \midrule
        Ours w/o CVAE & 97.27 & 78.52 & 91.80 & 91.80 & 84.38 & 96.48 & 97.66 \\
        Ours w/o Condition & 96.88 & 81.41 & 89.06 & 96.09 & 82.42 & 98.44 & 97.66 \\
        Ours w/o Foothold & 98.05 & 94.14 & 95.70 & 95.70 & 85.94 & 99.61 & 98.83 \\
        \bottomrule
    \end{tabular}
    }
\end{table}

We compare the distribution of policy-generated motions with MoCap data. For each terrain, we randomly sample consecutive 10-frame joint-position clips from each policy and compute the Maximum Mean Discrepancy (MMD)~\cite{MMD} against the corresponding MoCap clips. As shown in Figure~\ref{fig:mmd}, our method achieves the lowest average MMD of 0.1301 and consistently yields the lowest MMD across all terrains. T-GMP produces motion distributions closer to natural human MoCap than AMP and other baselines. We further report the mean and standard deviation of joint torques and accelerations across all terrains. As shown in Figure~\ref{fig:joint metrics}, our method yields an average joint torque of 181.2~$\mathrm{N\cdot m}$ and an average joint acceleration of 502.3~$\mathrm{rad/s^2}$, with standard deviations of 69.7~$\mathrm{N\cdot m}$ and 595.1~$\mathrm{rad/s^2}$. T-GMP exhibits lower torque and acceleration fluctuations, indicating smoother and more natural motions.

\subsection{Generalization and Robustness}
\label{sec:Generalization and Robustness}

\begin{figure}[t]
    \centering
    \includegraphics[width=1.0\textwidth]{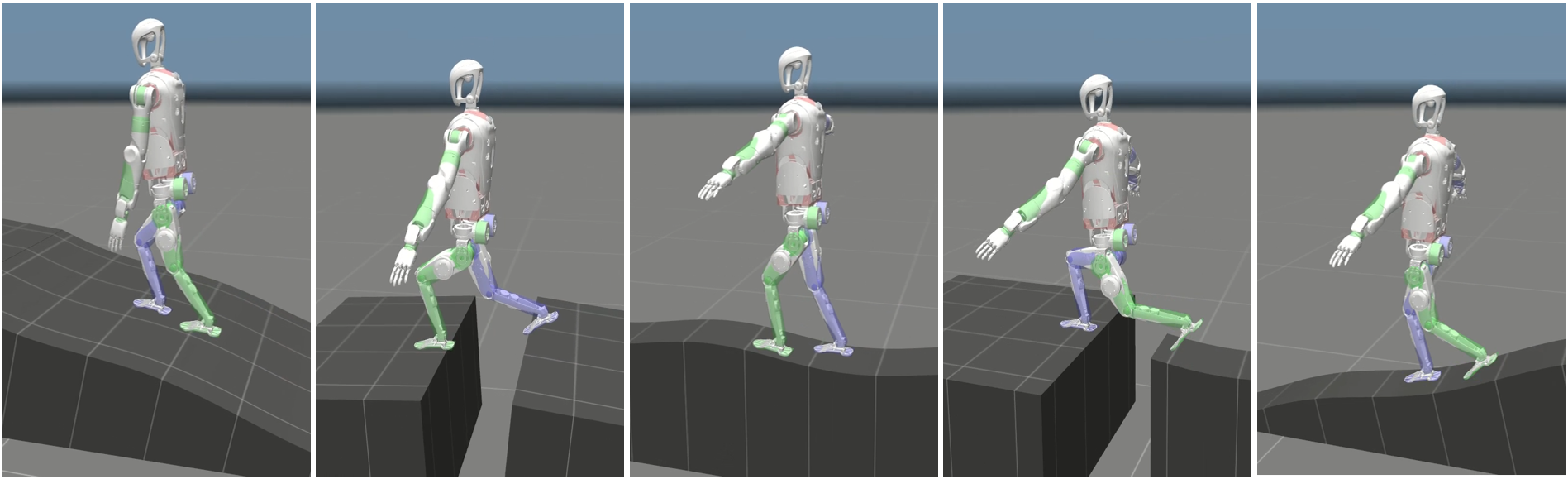}
    \caption{
    The robot traverses multiple unseen terrains in a single run, including an ascending wavy slope, a wavy balance beam, irregular gaps, and a descending wavy balance beam.
    }
    \label{fig:unseen terrain}
\end{figure}

As shown in Figure~\ref{fig:unseen terrain}, our method generalizes to multiple unseen terrains in MuJoCo and exhibits combinations of learned motion patterns (The robot combines arm extension for balance with \ac{com} lowering for slope adaptation when descending an unseen wavy beam), rather than simply replaying fixed motion patterns from the training data. We further evaluate the terrain traversal success rates of all methods across seven terrains, with detailed experimental settings provided in Section~\ref{success rate configuration} and results summarized in Table~\ref{tab:success rate}. Our method achieves the highest success rate on every evaluated terrain. In the beam traversal task, our method achieves an average improvement of 12.78 percentage points over the comparison methods, mainly due to the coordinated arm-extension behaviors for balance regulation. In the stair-descent task, our method achieves an average improvement of 9.58 percentage points, benefiting from adaptive center-of-mass lowering and the Foothold Penalty, which suppresses unstable edge contacts and sliding behaviors through reliable foothold control.

\subsection{Real-world Deployment}
\label{sec:real world deployment}

\begin{figure}[b]
    \centering
    \includegraphics[width=1.0\textwidth]{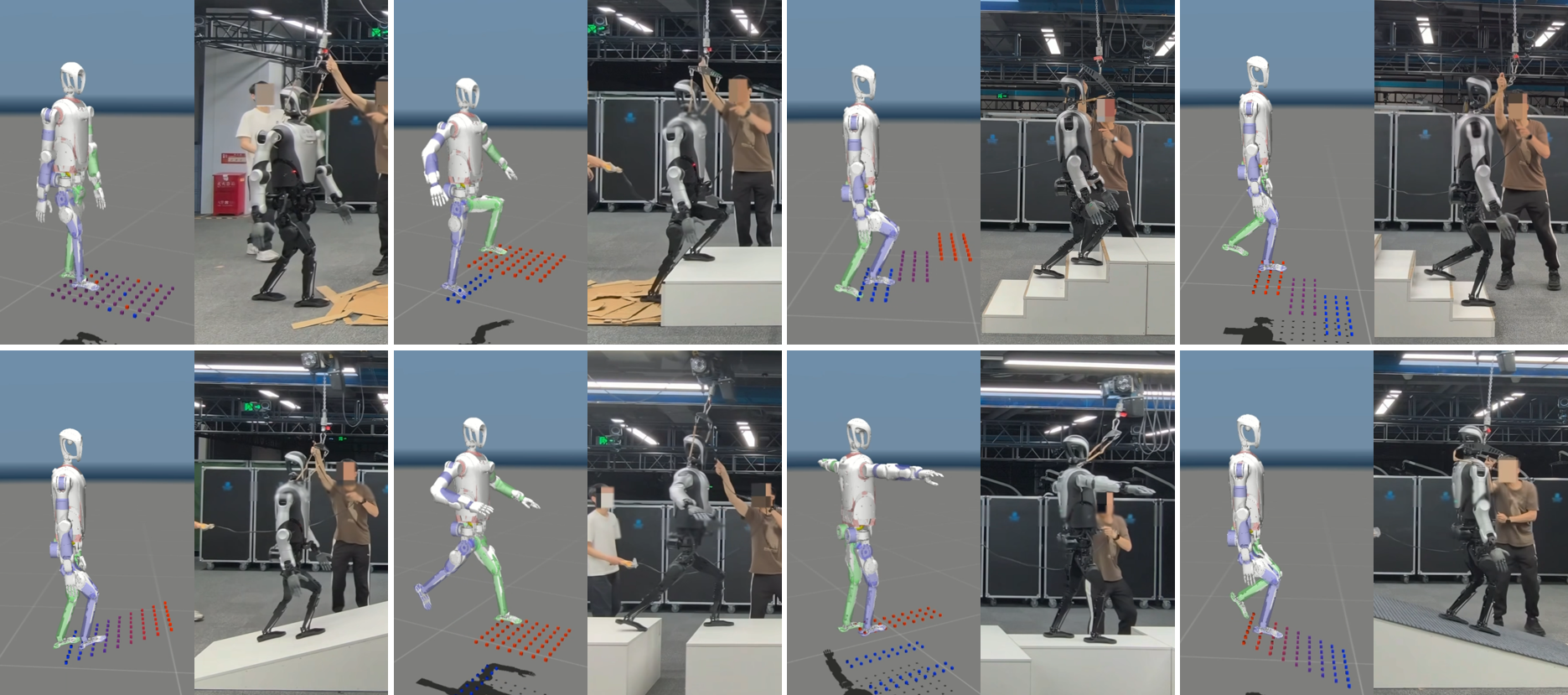}
    \caption{
    The robot exhibits terrain-adaptive and anthropomorphic locomotion behaviors across diverse terrains, including natural arm swinging during flat-ground walking, lowered center-of-mass postures on stairs and slopes, and arm extension for dynamic balance on gaps and beams.
    }
    \label{fig:all style}
\end{figure}

We compare expert demonstrations with real-robot deployments to analyze the consistency of terrain-conditioned motion distributions learned by the policy. Experiments are conducted on the full-size Kuavo humanoid robot over diverse terrains, including flat ground, a 0.3\,m high stage, stairs with 0.13\,m step height and 0.28\,m tread width, a 15$^\circ$ slope, a 0.4\,m wide gap, and a 0.35\,m wide beam (with the robot foot separation being 0.3\,m). As shown in Figure~\ref{fig:all style}, our method effectively preserves diverse terrain-conditioned expert motion patterns, while exhibiting coordinated and anthropomorphic locomotion behaviors during real-world deployment.

\section{Conclusion}
\label{sec:conclusion}

In this work, we propose T-GMP, which unifies anthropomorphic naturalness and robust terrain traversal for humanoid locomotion. By leveraging a CVAE, our method successfully learned a generative manifold of human movements that is intrinsically aligned with the terrain-aware physical constraints. Our approach moves beyond static motion priors by employing a terrain-conditioned discriminator that dynamically modulates the naturalness reward, enabling the agent to discover sophisticated balancing strategies, such as reflexive arm-spreading on a narrow beam. Experimental results demonstrate that T-GMP improves traversal robustness and motion smoothness by promoting physically coordinated and anthropomorphic whole-body behaviors, while producing diverse motion styles. Our work suggests that integrating generative motion priors conditioned on perceptual observations into \ac{rl} is a powerful paradigm for achieving the next level of agile and human-like robot locomotion.

\section{Limitations and Future Work}

First, T-GMP relies on paired motion states and local height maps for training, which limits its ability to directly leverage large-scale public motion datasets that lack terrain-aligned perceptual observations. Although a trained T-GMP can be reused across downstream policies, constructing such paired data still requires additional effort. Future work could follow~\cite{EgoHTR} to directly collect diverse \ac{mocap} data aligned with perceptual observations, thereby simplifying the data construction pipeline.

Second, the proposed terrain perception pipeline relies on LiDAR-based environment reconstruction. However, LiDAR measurements inevitably contain sensing noise and reconstruction artifacts. Such perception uncertainties may compromise locomotion robustness on complex terrains. Meanwhile, terrain reconstruction suffers from limited update frequency. Future work may investigate multi-modal perception strategies that integrate depth cameras with LiDAR sensing to improve reconstruction accuracy and perception update frequency.




\clearpage
\bibliography{ref}  

\clearpage
\appendix

\section{Data Collection}
\label{app:data collection}
\subsection{Privileged Policy Training}

\begin{figure}[h]
    \centering
    \includegraphics[width=1.0\textwidth]{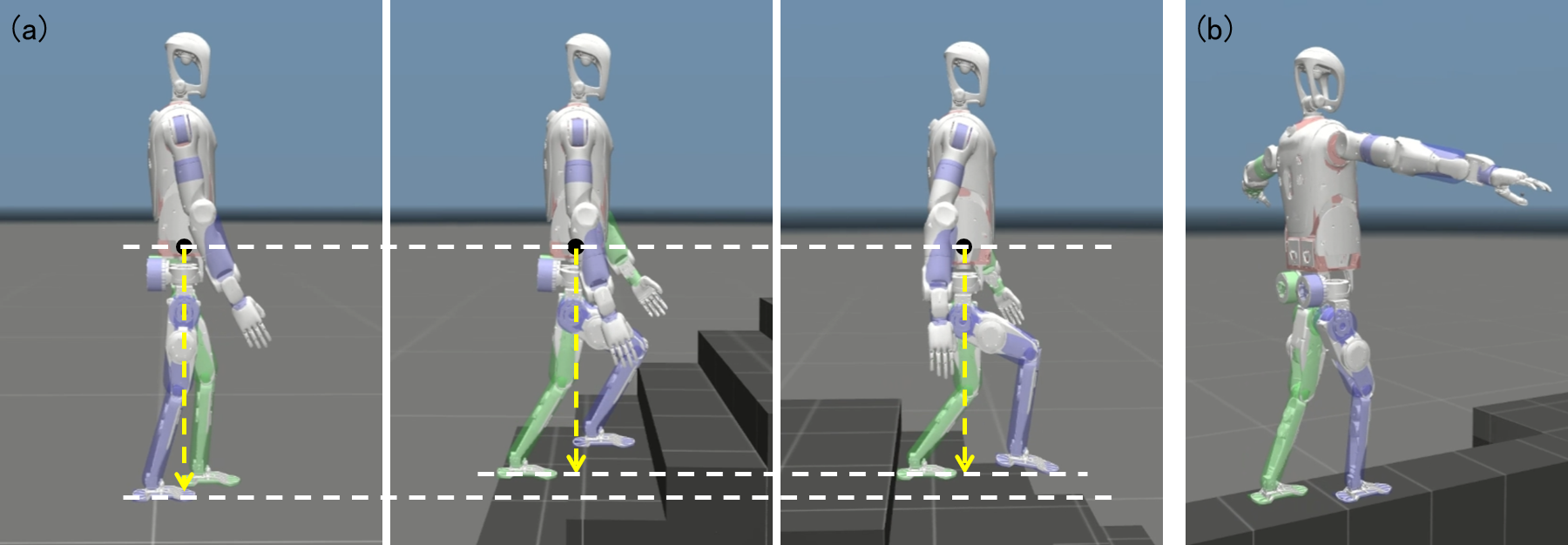}
    \caption{
    For challenging terrains such as stairs and beams, we introduce terrain-specific constraints on (a) center-of-mass regulation and (b) arm-balancing behaviors to promote more anthropomorphic and robust whole-body locomotion.
    }
    \label{fig:privileged policy}
\end{figure}

Each privileged policy $\pi_i^{\text{priv}}$ is trained exclusively on terrain $i$, using the corresponding MoCap dataset $\mathcal D_i^{\text{mocap}}$ as the AMP~\cite{amp} motion prior to learn anthropomorphic and physically feasible behaviors. Unlike the downstream policy, $\pi_i^{\text{priv}}$ receives the privileged observations listed in Table~\ref{tab:observation terms} as additional inputs, which are unavailable during real-world deployment. The privileged observations accelerate training, allowing $\pi_i^{\text{priv}}$ to be trained in $\sim$12 hours with 2048 parallel environments on a single RTX 4090. Figure~\ref{fig:privileged policy} illustrates the distinct terrain-specific behaviors learned by the privileged policies. Additionally, CVAE training takes only $\sim$3 hours. Therefore, extending T-GMP to a new terrain $j$ requires only collecting the corresponding MoCap data, training $\pi_j^{\text{priv}}$, and retraining the CVAE with the augmented dataset, which can typically be completed within one day.

\subsection{Data Collection Details}

\begin{wrapfigure}[18]{r}{0.42\textwidth}
    \vspace{-12pt}
    \centering
    \includegraphics[width=\linewidth]{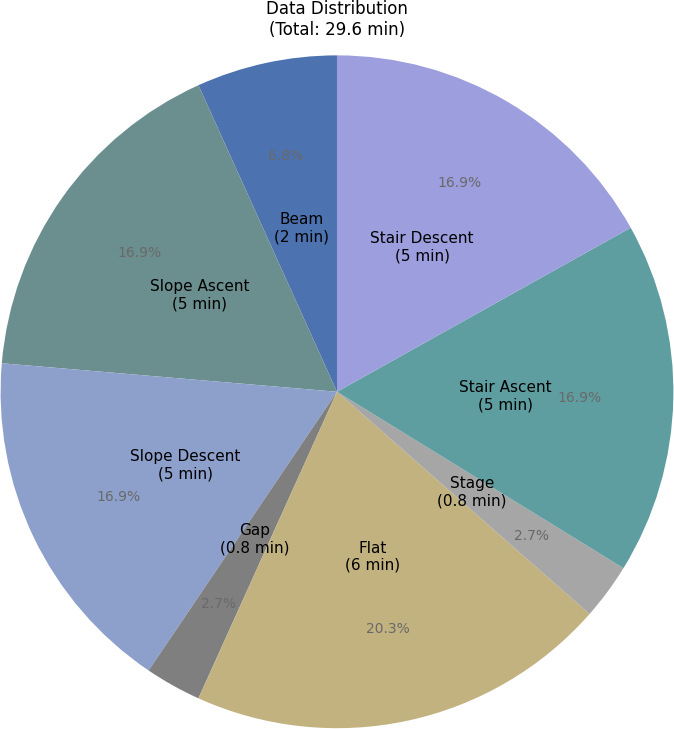}
    \caption{
    Distribution of collected motion data across eight terrains.
    }
    \label{fig:data efficiency}
\end{wrapfigure}

Table~\ref{tab:terrain settings} summarizes the detailed settings used for data collection. Our settings cover common configurations within each terrain category. T-GMP can interpolate between neighboring terrain configurations and generalize to unseen intermediate parameters. For example, motion data collected at step heights of 0.1\,m and 0.15\,m on stairs enable T-GMP to generate an appropriate motion style for a step height of 0.13\,m. This generalization capability is further evidenced by the composition of learned motion styles on unseen terrains, as shown in Figure~\ref{fig:unseen terrain}. Benefiting from the proposed T-GMP, we learn a terrain-conditioned motion manifold spanning eight terrains using only approximately 29.6\,min (88.8k frames) of data. Figure~\ref{fig:data efficiency} reports the duration and proportion of data collected for each terrain. T-GMP effectively captures diverse anthropomorphic locomotion patterns from a small amount of expert demonstrations while covering diverse terrain conditions, locomotion behaviors, and terrain difficulty levels, demonstrating strong data efficiency and motion generalization capabilities.

\begin{table}[t]
    \centering
    \caption{Data collection settings.}
    \label{tab:terrain settings}
    \renewcommand{\arraystretch}{1.2}
    \begin{tabular}{l|ccccc}
        \toprule
        Terrain & \multicolumn{5}{c}{Settings} \\
        \midrule
        Flat          & -- & Forward & Backward & Turn Left & Turn Right \\
        Gap           & Width & 0.1\,m & 0.2\,m & 0.3\,m & 0.4\,m \\
        Beam          & Width & 0.2\,m & 0.3\,m & 0.4\,m & -- \\
        Stage         & Height & 0.1\,m & 0.2\,m & 0.3\,m & -- \\
        Stair Ascent  & Step Height & 0.05\,m & 0.1\,m
        & 0.15\,m & 0.2\,m \\
        Stair Descent & Step Height & 0.05\,m & 0.1\,m 
        & 0.15\,m & 0.2\,m \\
        Slope Ascent  & Tilt Angle & $5^\circ$  & $10^\circ$ & $15^\circ$ & $20^\circ$ \\
        Slope Descent & Tilt Angle & $5^\circ$  & $10^\circ$ & $15^\circ$ & $20^\circ$ \\
        \bottomrule
    \end{tabular}
\end{table}

\section{Policy Training}

\subsection{Observations}
\label{app:observations}

Table~\ref{tab:observation terms} summarizes all observations used by the Actor and Critic. The Actor inputs consist of command observations, perceptual observations, and policy observations, enabling the policy to jointly capture robot dynamics and local terrain geometry. Specifically, the velocity command $c_t$ consists of linear and angular velocity commands, i.e., $c_t=[v_t,~\omega_t]$. To improve value estimation accuracy, we introduce privileged observations that are unavailable during real-world deployment into the Critic network during training. These privileged observations are only used to facilitate more accurate value learning for the Critic and are not provided to the Actor during policy execution, thereby ensuring deployment compatibility and perception consistency under real-world conditions.

\begin{table}[h]
    \centering
    \caption{Observation Terms}
    \label{tab:observation terms}
    \renewcommand{\arraystretch}{1.2}
    \begin{tabular}{@{}lccc@{}}
        \toprule
        \textbf{Term} & \textbf{Notation} & \textbf{Type} & \textbf{Description} \\
        \midrule
        Velocity command & $c_t$ & Command & Velocity command of the robot's base frame \\
        Height scan & $h_t$ & Perception & A $9 \times 7$ height scan in front of the robot \\
        Base angular velocity & $\omega_t$ & Policy & Angular velocity in the robot's base frame \\
        Projected gravity & $\mathbf g_t$ & Policy & Gravity projection on the robot's base frame \\
        Joint position & $q_t$ & Policy & The joint positions of the robot \\
        Joint velocity & $\dot q_t$ & Policy & The joint velocities of the robot \\
        Last action & $a_{t-1}$ & Policy & The last input action to the environment \\
        Base linear velocity & $v_t$ & Privileged & Linear velocity in the robot's base frame \\
        Joint torque & $\tau_t$ & Privileged & Torques applied to the robot's joints \\
        Joint acceleration & $\ddot q_t$ & Privileged & The joint accelerations of the robot \\
        Feet linear velocity & $v^{\text{feet}}_t$ & Privileged & The linear velocity of the robot's feet \\
        Feet contact force & $F^{\text{contact}}_t$ & Privileged & Contact force on the robot's feet \\
        Mass & $m$ & Privileged & Total mass of the robot \\
        Material & $\mu$ & Privileged & Friction coefficient of the ground \\
        Center of mass & $p_{\text{com}}$ & Privileged & Center of mass in the robot's base frame \\
        Action delay & $t_{\text{delay}}$ & Privileged & Action delay on the robot's motors \\
        Push force & $F^{\text{ext}}_t$ & Privileged & External push force applied on the robot \\
        Push torque & $\tau^{\text{ext}}_t$ & Privileged & External push torque applied on the robot \\
        Feet height & $h^{\text{feet}}_t$ & Privileged & Robot's feet height \\
        Feet air time & $t_{\text{air}}$ & Privileged & Robot's feet air time since last contact \\
        \bottomrule
    \end{tabular}
\end{table}

\begin{table}[h]
    \centering
    \caption{PPO Hyperparameters}
    \label{tab:ppo hyperparameters}
    \renewcommand{\arraystretch}{1.2}
    \begin{tabular}{@{}lc@{}}
        \toprule
        \textbf{Parameter} & \textbf{Value} \\
        \midrule
        Batch size & $24 \cdot 2048 = 49152$ \\
        Mini batch size & $6 \cdot 2048 = 12288$ \\
        Number of epochs & $5$ \\
        Clip range & $0.2$ \\
        Entropy coefficient & $0.005$ \\
        Discount factor & $0.99$ \\
        GAE discount factor & $0.95$ \\
        Desired KL-divergence & $0.01$ \\
        Learning rate &  $0.001$\\
        Max gradient norm & $1.0$ \\
        \bottomrule
    \end{tabular}
\end{table}

\subsection{Reinforcement Learning Setup}
\label{app:rl learning setup}

Policy optimization is performed using \ac{ppo}~\cite{ppo}, with the associated training hyperparameters summarized in Table~\ref{tab:ppo hyperparameters}. All experiments are conducted in the Isaac Lab~\cite{isaaclab} simulation environment on a single NVIDIA RTX 4090 GPU, with 2048 environments running in parallel. We adopt an asymmetric Actor-Critic architecture to train the policy, where both the Actor and Critic are parameterized by multilayer perceptrons (MLPs) with hidden layer sizes of [512, 256, 128]. The discriminator is implemented as an MLP with hidden layer sizes of [1024, 512, 256], followed by a linear output layer that maps the final hidden representation to a scalar score.

\subsection{Reward Formulation}
\label{app:reward formulation}

\begin{table}[b]
    \centering
    \caption{Reward Terms}
    \label{tab:reward terms}
    \renewcommand{\arraystretch}{1.2}
    \begin{tabular}{@{}lcc@{}}
        \toprule
        \textbf{Term} & \textbf{Function} & \textbf{Weight} \\
        \midrule
        Linear velocity command tracking & $\exp (-\|v_{xy}-v_{xy}^{\text{cmd}}\|^2)$ & $5.0$ \\
        Angular velocity command tracking & $\exp (-\|\omega_{z}-\omega_{z}^{\text{cmd}}\|^2)$ & $3.0$ \\
        \midrule
        Joint velocity & $\|\dot q\|^2$ & $-2.0\times10^{-3}$ \\
        Joint acceleration & $\|\ddot q\|^2$ & $-2.5\times10^{-7}$ \\
        Joint torque & $\|\tau\|^2$ & $-1.0\times10^{-5}$ \\
        Joint power & $|\tau\cdot\dot q|$ & $-2.0\times10^{-5}$ \\
        Action rate & $\|a_t-a_{t-1}\|^2$ & $-0.005$ \\
        Action smoothness & $\|(a_t-a_{t-1})-(a_{t-1}-a_{t-2})\|^2$ & $-0.01$ \\
        Joint limit & $\max(0,q-q_{\max})+\max(0,q_{\min}-q)$ & $-1.0$ \\
        Feet slide & $\|v^{\text{feet}}_{xy}\|^2_2*\mathbb I_{\text{contact}}$ & $-0.1$ \\
        Contact force & $\max(0,f_{\text{contact}}-f_\text{{threshold}})$ & $-0.001$ \\
        \midrule
        Termination & $\mathbb I_{\text{reset}}$ & $-200$ \\
        \bottomrule
    \end{tabular}
\end{table}

Since motion style is primarily guided by the discriminator, the reward design remains concise yet effective. The reward received by the policy during training consists of four components: (1) task reward $r_\text{task}$, which encourages the robot’s root to track the commanded linear and angular velocities, (2) regularization reward $r_\text{reg}$, which enforces smoothness in actions and states while ensuring physical plausibility, (3) style reward $r_\text{amp}$, provided by the style discriminator, which encourages policy-induced state transitions to align with the expert motion styles, and (4) foothold penalty $r_{\text{foothold}}$, which penalizes undesirable foot-contact behaviors. The overall training reward is defined as:
\begin{equation}
r=w_1r_\text{task}+w_2r_\text{reg}+w_3r_\text{amp}+w_4r_{\text{foothold}},
\end{equation}
where $w_i$ represents the corresponding weights.

Benefiting from the inclusion of the style reward, the policy becomes significantly less sensitive to reward weighting, leading to a more stable and efficient hyperparameter tuning process. In practice, a single set of reward parameters can be reused across subsequent training runs with only minimal adjustment. We design a unified and concise reward function to train all policies. We primarily focus on the smoothness of robot motion and energy consumption, enabling the reward function to remain consistent across different tasks and experimental settings, thereby avoiding additional task- or scenario-specific reward engineering. The definitions of all reward terms and their corresponding weights are summarized in Table~\ref{tab:reward terms}.

\subsection{Foothold Penalty}

\begin{figure}[t]
    \centering
    \begin{subfigure}[b]{0.495\textwidth}
        \centering
        \includegraphics[width=\textwidth]{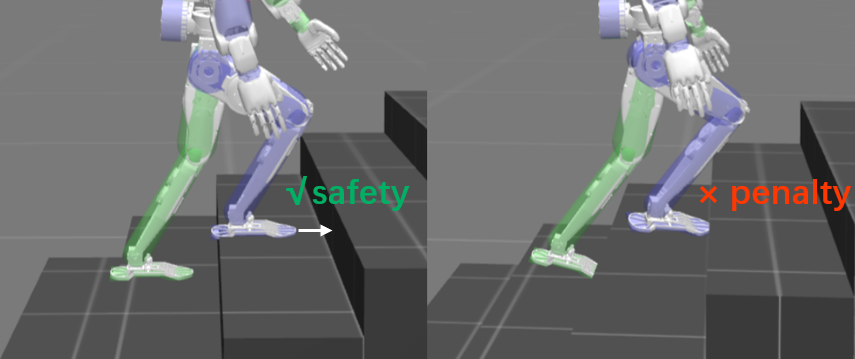}
        \caption{
        }
        \label{fig:toe penalty}
    \end{subfigure}
    \hfill
    \begin{subfigure}[b]{0.495\textwidth}
        \centering
        \includegraphics[width=\textwidth]{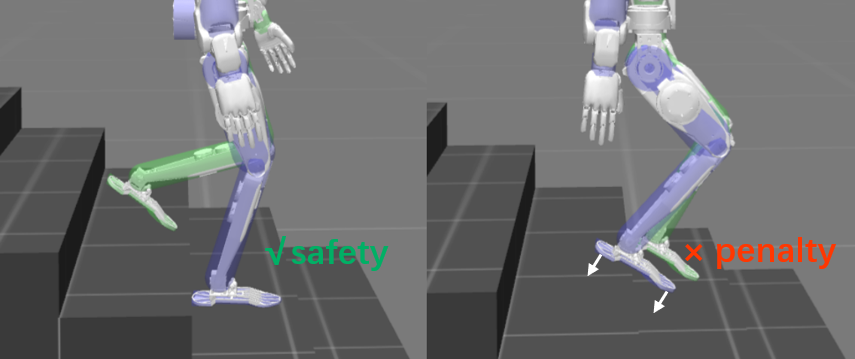}
        \caption{
        }
        \label{fig:sole penalty}
    \end{subfigure}
    \caption{
    Illustration of the Foothold Penalty mechanism. (a) During stair ascent, the minimum toe-terrain contact distance $d_{\text{toe}}$ is evaluated to reduce potential toe-collision risks. (b) During stair descent, the maximum sole-terrain contact distance $d_{\text{sole}}$ is evaluated to suppress unstable foothold behaviors such as edge contacts and slipping.
    }
    \label{fig:foothold penalty}
\end{figure}

We design a Foothold Penalty mechanism to explicitly regularize the contact quality between the robot feet and the terrain. As illustrated in Figure~\ref{fig:toe penalty}, during terrain traversal, we compute the minimum distance $d_{\text{toe}}$ between the toe and the terrain surface. A penalty is applied when the toe–surface distance is below the safety threshold $d_{\text{toe}}^{\text{threshold}}$, suppressing potential toe-collision behaviors. Meanwhile, as illustrated in Figure~\ref{fig:sole penalty}, we further compute the maximum distance $d_{\text{sole}}$ between the sole and the terrain surface. When the distance exceeds the threshold $d_{\text{sole}}^{\text{threshold}}$, an additional penalty is imposed to discourage unstable support behaviors such as edge contacts and slipping. This mechanism encourages the policy to learn more stable and physically feasible foothold patterns, thereby improving locomotion stability and traversal robustness on complex terrains.

\section{Success Rate Configurations}
\label{success rate configuration}

We evaluate terrain traversal success rates on seven representative challenging terrains, including a 0.3\,m high stage, stairs with 0.13\,m step height and 0.28\,m tread width, a 15$^\circ$ slope, a 0.4\,m wide gap, and a 0.35\,m wide beam. Each terrain is constructed within an 8\,m$\times$8\,m evaluation area. A trial is considered successful if the robot completes a full terrain traversal within a predefined time limit. Failure is recorded if any body part other than the feet collides with the terrain during traversal, any joint exceeds its position limits or exhibits excessively large accelerations, or if the robot fails to complete the traversal within the allotted time.

\section{Sim-to-Real Details}
\subsection{Robot Setup}

The full-size Kuavo humanoid robot is 1.66\,m tall, weighs approximately 55\,kg, and possesses 28 \ac{dofs}, including 2 \ac{dofs} in the head, 14 \ac{dofs} in the arms, and 12 \ac{dofs} in the legs. The robot has 28 \ac{dofs}, while the policy controls 26 actuated joints, excluding the two head joints. A Mid-360 LiDAR is mounted on the robot head for point-cloud perception and terrain reconstruction. The system adopts a dual-computing architecture, where an Orin-NX module processes point-cloud perception data in real time, while a MoFang i9-13900 computer executes low-level whole-body locomotion control.

The robot is controlled using a joint-space PD controller with control torques defined as:
\begin{equation}
\tau = K_\text{p}(q^*-q)-K_\text{d}\dot q,
\end{equation}
where $K_\text{p}$ and $K_\text{d}$ denote proportional and derivative gains, respectively, $q$ denotes the current joint position, and $\dot q$ denotes the current joint velocity. The policy action $a$ is defined as a residual control signal relative to the default joint configuration $q_0$, such that the target joint position is given by
\begin{equation}
q^*=q_0+a.
\end{equation}

\subsection{Acquisition of Elevation Map}

\begin{figure}[t]
    \centering
    \begin{subfigure}[b]{0.495\textwidth}
        \centering
        \includegraphics[width=\textwidth]{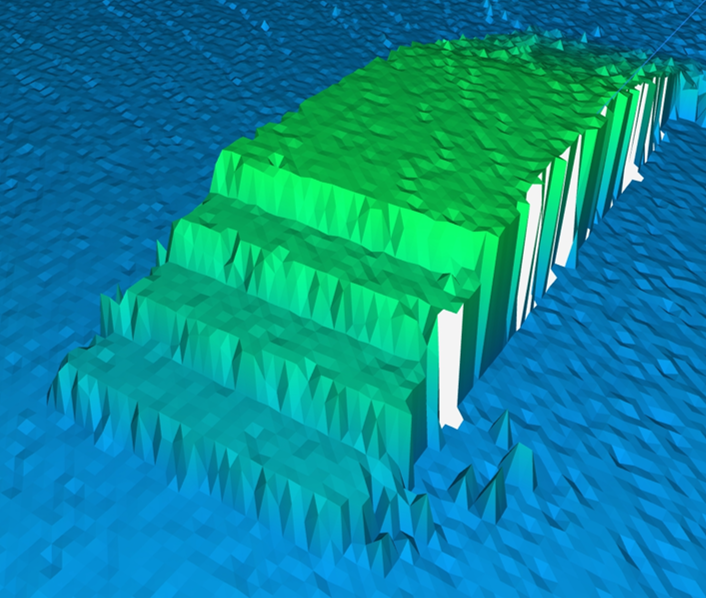}
        \caption{
        }
        \label{fig:elevation raw}
    \end{subfigure}
    \hfill
    \begin{subfigure}[b]{0.495\textwidth}
        \centering
        \includegraphics[width=\textwidth]{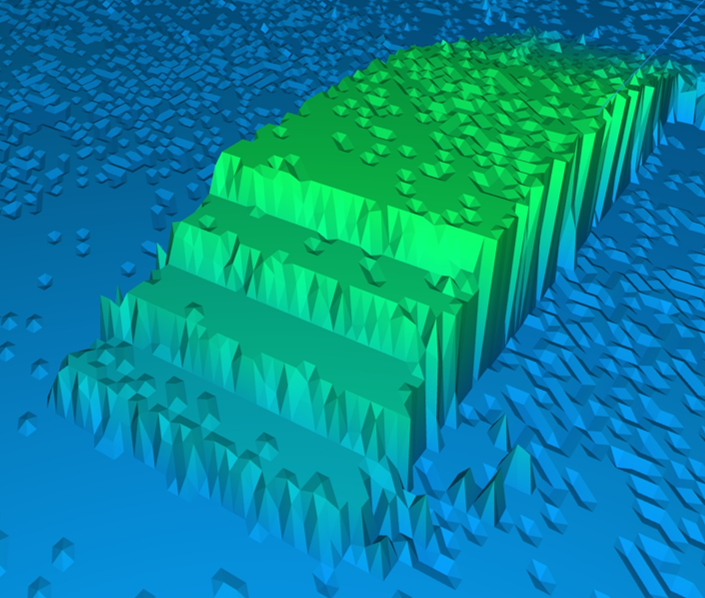}
        \caption{
        }
        \label{fig:elevation inpaint}
    \end{subfigure}
    \caption{
    Elevation map construction pipeline. (a) Raw dense elevation map generated from sparse environmental point clouds, containing missing regions caused by incomplete point-cloud reconstruction. (b) Processed elevation map after applying the nearest-region filling strategy, which is used for subsequent height-map sampling.
    }
    \label{fig:elevation map processing}
\end{figure}

As illustrated in Figure~\ref{fig:elevation raw}, following the elevation mapping pipeline in~\cite{elevation}, the raw point cloud observations are converted into a dense elevation map representation. To avoid invalid \texttt{NaN} inputs caused by sparse point-cloud reconstruction, we apply a nearest-region filling strategy that replaces \texttt{NaN} cells with the minimum height value within the surrounding 5$\times$5 neighborhood, resulting in the processed elevation map shown in Figure~\ref{fig:elevation inpaint}. We then uniformly sample a local terrain patch of size 0.8\,m$\times$0.6\,m in front of the robot from the processed elevation map, which is used as the terrain perception input to the policy.

\section{Motion Style Visualization}
\label{sec:Motion Style Visualization}

\begin{figure}[t]
    \centering
    \includegraphics[width=1.0\textwidth]{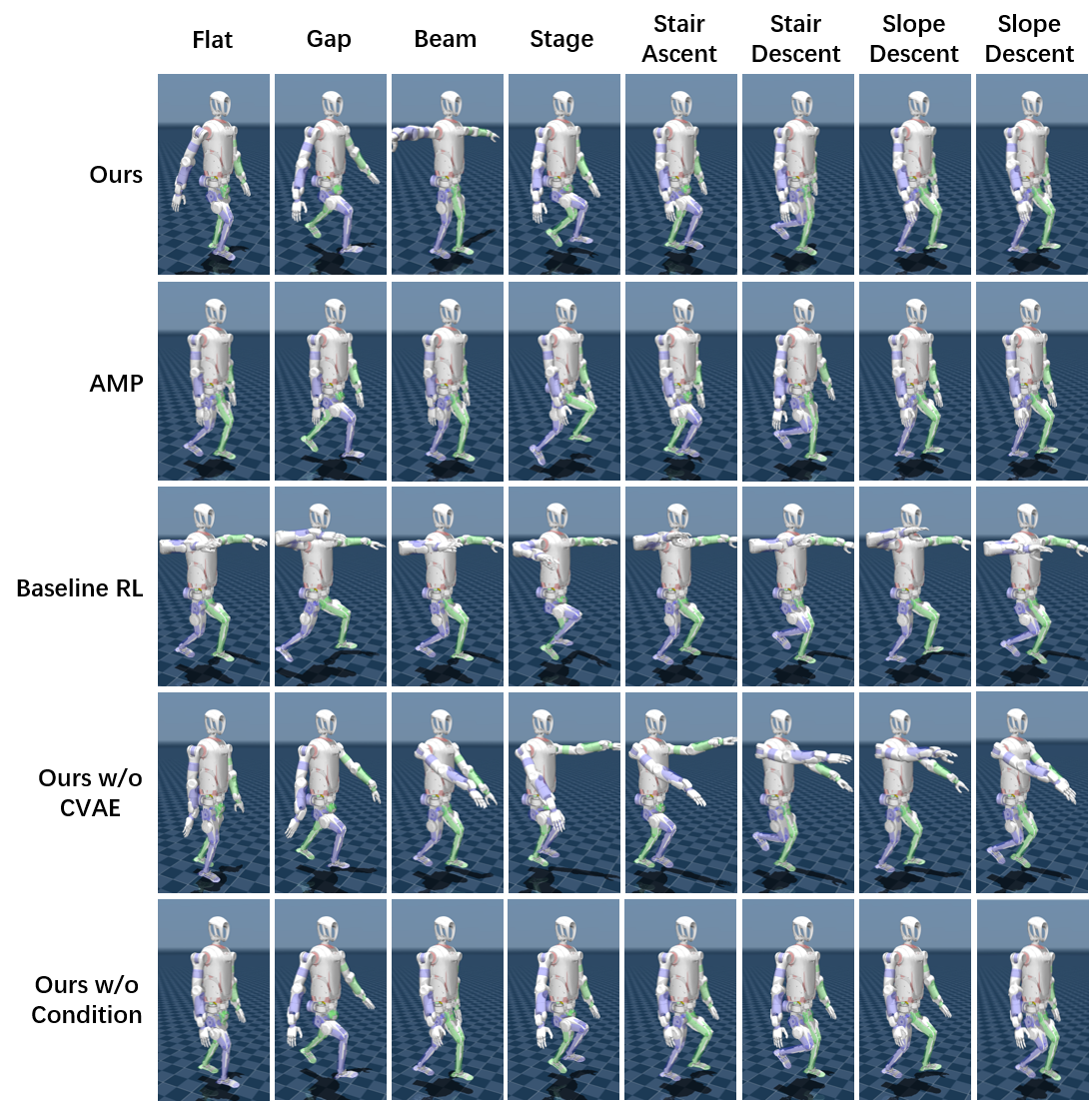}
    \caption{
    Motion styles produced by different policies across all terrains.
    }
    \label{fig:policy style}
\end{figure}

As shown in Figure~\ref{fig:policy style}, we compare the motion styles produced by different policies during terrain traversal. Baseline RL is trained without explicit style guidance or additional reward terms for shaping upper-body motion. Thus, it produces uniform yet uncoordinated motions across terrains. Ours w/o CVAE employs a terrain-conditioned discriminator, while its expert state transitions are randomly sampled from the expert dataset $\mathcal D$ and may therefore be inconsistent with the current terrain condition. Such mismatched terrain–state pairs hinder discriminator convergence, resulting in incoherent mixtures of motion patterns.

\end{document}